\documentclass[conference]{IEEEtran}
\IEEEoverridecommandlockouts

\usepackage{cite}
\usepackage{amsmath,amssymb,amsfonts}
\usepackage{algorithm, algorithmic}
\usepackage{graphicx}
\usepackage{textcomp}
\usepackage{xcolor}
\usepackage{hyperref}
\usepackage{cleveref}
\usepackage{tikz}
\usepackage{bbm}
\usetikzlibrary{decorations.pathreplacing, shapes.arrows}
\usetikzlibrary{positioning}
\usepackage{pgfplots}
\usepackage{pgfplotstable}
\pgfplotsset{compat=1.14}
\usepgfplotslibrary{statistics}
\usepackage{csvsimple}
\usepackage{multirow}
\usepackage[inline]{enumitem}
\usepackage[normalem]{ulem}
\usetikzlibrary{shapes,arrows,positioning}
\def\BibTeX{{\rm B\kern-.05em{\sc i\kern-.025em b}\kern-.08em
		T\kern-.1667em\lower.7ex\hbox{E}\kern-.125emX}}

\DeclareMathOperator*{\argmin}{arg\,min}
\DeclareMathOperator*{\argmax}{arg\,max}

\allowdisplaybreaks

\begin{document}
\title{Shrinking POMCP: A Framework for Real-Time UAV Search and Rescue\\
}

\author{
\IEEEauthorblockN{
Yunuo Zhang\IEEEauthorrefmark{1},
Baiting Luo\IEEEauthorrefmark{1}, 
Ayan Mukhopadhyay\IEEEauthorrefmark{1}, 
Daniel Stojcsics\IEEEauthorrefmark{1},
Daniel Elenius\IEEEauthorrefmark{2}, \\
Anirban Roy\IEEEauthorrefmark{2}, 
Susmit Jha\IEEEauthorrefmark{2}, 
Miklos Maroti\IEEEauthorrefmark{3}, 
Xenofon Koutsoukos\IEEEauthorrefmark{1}, 
Gabor Karsai\IEEEauthorrefmark{1} and 
Abhishek Dubey\IEEEauthorrefmark{1}
}
\IEEEauthorblockA{
\IEEEauthorrefmark{1}Vanderbilt University, \IEEEauthorrefmark{2}SRI, \IEEEauthorrefmark{3}University of Szeged
}
}

\maketitle

\begin{abstract}
Efficient path optimization for drones in search and rescue operations faces challenges, including limited visibility, time constraints, and complex information gathering in urban environments. We present a comprehensive approach to optimize UAV-based search and rescue operations in neighborhood areas, utilizing both a 3D AirSim-ROS2 simulator and a 2D simulator. The path planning problem is formulated as a partially observable Markov decision process (POMDP), and we propose a novel ``Shrinking POMCP'' approach to address time constraints. In the AirSim environment, we integrate our approach with a probabilistic world model for belief maintenance and a neurosymbolic navigator for obstacle avoidance. The 2D simulator employs surrogate ROS2 nodes with equivalent functionality. We compare trajectories generated by different approaches in the 2D simulator and evaluate performance across various belief types in the 3D AirSim-ROS simulator. Experimental results from both simulators demonstrate that our proposed shrinking POMCP solution achieves significant improvements in search times compared to alternative methods, showcasing its potential for enhancing the efficiency of UAV-assisted search and rescue operations.
\end{abstract}


\section{Introduction} \label{sec:introduction}
{\setlength{\parskip}{0.1em}
Search and rescue (SAR) operations are critical, time-sensitive missions conducted in challenging environments like neighborhoods, wilderness \cite{wilderness_search}, or maritime settings \cite{Maritime_search}. These resource-intensive operations require efficient path planning and optimal routing \cite{UAV_Path_Planning_Lin_2009}. In recent years, Unmanned Aerial Vehicles (UAVs) have become valuable SAR assets, offering advantages such as rapid deployment, extended flight times, and access to hard-to-reach areas. Equipped with sensors and cameras, UAVs can detect heat signatures, identify objects, and provide real-time aerial imagery to search teams \cite{UAV_Search_Waharte_2010}.

However, the use of UAVs in SAR operations presents unique challenges, particularly in path planning and decision-making under uncertainty. Factors such as limited battery life, changing weather conditions, and incomplete information about the search area complicate the task of efficiently coordinating UAV movements to maximize the probability of locating targets \cite{UAV_Path_Planning_Lin_2009}. To address these challenges, researchers have proposed formalizing UAV path planning for SAR missions as partially observable Markov decision processes (POMDPs)~\cite{UAV_POMDPs_Trotti_2023,UAV_POMDPs_Ahmadi_2021,UAV_POMDPs_Floriano_2021}. POMDPs provide a mathematical framework for modeling sequential decision-making problems in uncertain environments where the system's state is not fully observable \cite{POMDPs_Kaelbling_1998}. 

POMDP-like planning is crucial for search operations due to inherent uncertainties \cite{Planning_Roy_2005}. In UAV-based SAR, POMDPs capture uncertainties in target locations, sensor observations, and environmental conditions while optimizing UAV paths \cite{UAV_Search_Waharte_2010}. They model unknown environmental states, imperfect sensor information \cite{Perception_Spaan_2008}, and the complex interdependence between decisions and future observations \cite{SARSOP}. POMDPs naturally address partial observability and long-term action consequences \cite{Perception_Spaan_2008}. However, solving large-scale POMDP problems remains computationally challenging, with complexity growing exponentially with state space, observation space, and planning horizon sizes, often making exact solutions intractable for real-world applications \cite{POMDPs_Complexity_Papadimitriou_1987}.

To address this challenge, recent research has focused on online POMDP solutions, aiming to find good policies quickly by interleaving planning and execution and using sampling-based techniques to explore the belief space efficiently \cite{POMCP_Silver_2010, DESPOT_Somani_2013}. Online POMDP frameworks have been applied to UAV path planning for SAR operations, addressing uncertainties in target motion and sensor observations \cite{POMDPs_UAV_Ragi_2013}, partial observability of victim locations and environmental hazards \cite{POMDPs_UAV_Carpin_2013}, and challenges in multi-UAV search missions \cite{POMDPs_UAV_Carabaza_2017}. Despite these advancements, computational efficiency under strict time constraints remains a critical challenge for real-time applications. 

This paper presents a novel online path planner for UAVs designed to enhance the efficiency of search and rescue operations in urban environments. Our approach combines advanced simulation techniques with an innovative POMDP formulation and solution approach. This method, called Shrinking POMCP (partially observable Monte Carlo planning), guides the agent toward the next best non-sparse region for planning (we define a sparse region as a region that has probability of target appearing in that region less than a given threshold). This innovation is particularly crucial for real-world applications with strict time constraints, as it allows for more effective decision-making within limited computational resources. We demonstrate the effectiveness of the approach using an Airsim-based simulator.

The outline of this paper is as follows. We first describe the necessary background concepts (\S \ref{sec:background}), followed by the problem formulation (\S \ref{sec:problem}), solution framework (\S \ref{sec:framework}) and description of the POMDP planning algorithm (\S \ref{sec:planner}), the primary contribution of this paper. We conclude the paper with a description of metrics and experimental results (\S \ref{sec:analysis}).
}

\section{Background and related research} \label{sec:background}
POMDPs, or Partially Observable Markov Decision Processes, are a mathematical framework for modeling decision-making in situations where an agent must make decisions in an environment that is not fully observable. A POMDP is formally defined as a tuple $(\mathcal{S}, \mathcal{A}, \mathcal{T}, \mathcal{R}, \mathcal{O}, \mathcal{Z})$, where:

\begin{itemize}
	\item $\mathcal{S}$ is a finite set of states.
	\item $\mathcal{A}$ is a finite set of actions.
	\item $\mathcal{T}: \mathcal{S} \times \mathcal{A} \times \mathcal{S} \rightarrow [0,1]$ is the transition function, where $\mathcal{T}(s'|s,a)$ gives the probability of transitioning to state $s'$ given action $a$ is taken in state $s$.
	\item $\mathcal{R}: \mathcal{S} \times \mathcal{A} \rightarrow \mathbb{R}$ is the reward function.
	\item $\mathcal{O}$ is a finite set of observations.
	\item $\mathcal{Z}: \mathcal{S} \times \mathcal{A} \times \mathcal{O} \rightarrow [0,1]$ is the observation function, where $\mathcal{Z}(o|s',a)$ is the probability of observing $o$ given action $a$ is taken and the system transitions to state $s'$.
\end{itemize}

In a POMDP, the agent maintains a belief state $b(s)$, which is a probability distribution over all possible states. The belief state is updated after each action and observation using Bayes' rule. Solving a POMDP involves finding an optimal policy $\pi^*$ that maximizes the expected cumulative reward. However, exact solutions to POMDPs are computationally intractable for all but the smallest problems. As a result, much research has focused on approximate methods, including Point-based Value Iteration (PBVI) \cite{PBVI_Pineau_2003}, Heuristic Search Value Iteration (HSVI) \cite{HSVI_Smith_2012}, and Monte Carlo Tree Search (MCTS) methods, such as POMCP \cite{POMCP_Silver_2010}. Note that Partially Observable Monte Carlo Planning (POMCP) is an online POMDP solver that extends Monte Carlo Tree Search (MCTS) to POMDPs \cite{POMCP_Silver_2010}. (MCTS) is a heuristic search algorithm for decision processes, particularly effective in large state spaces.

Key components of MCTS include:
\begin{enumerate*}
	\item \textbf{Selection:} Starting from the root node, a child selection policy is recursively applied to descend through the tree until reaching a leaf node.
	\item \textbf{Expansion:} If the leaf node is not terminal and is within the computational budget, one or more child nodes are added to expand the tree.
	\item \textbf{Simulation:} A simulation is run from the new node(s) according to the default policy to produce an outcome.
	\item \textbf{Backpropagation:} The simulation result is then `backed up' through the selected nodes to update their statistics.
\end{enumerate*}

During selection phase, UCT (Upper Confidence Bounds for Trees) \cite{Bandit_Kocsis_2006} is used as child selection policy: $UCT = \overline{X_j} + C\sqrt{\frac{2\ln n}{n_j}}$, Where $\overline{X_j}$ is the average reward from node $j$, $n$ is the number of times the parent node has been visited, $n_j$ is the number of times child $j$ has been visited, and $C$ is an exploration constant.

Saisubramanian et al. \cite{GUSSP} introduced the Goal Uncertain Stochastic Shortest Path (GUSSP) problem, a specialized form of POMDP. GUSSPs extend the Stochastic Shortest Path framework to handle goal uncertainty, maintaining a belief state over possible goal configurations. While including an observation function for goals like POMDPs, GUSSPs simplify the problem by assuming full current state observability and myopic goal observations, resulting in a more tractable solution space compared to general POMDPs.

Despite advancements, existing approaches to solving POMDPs and GUSSPs face challenges in real-time applications. The computational complexity often exceeds time constraints of real-world scenarios. Even with GUSSP simplifications, the problem remains computationally demanding for large state spaces. This highlights the need for efficient algorithms that can provide quality solutions within tight time bounds, especially for robotics and autonomous systems requiring rapid decisions.

\subsection{AirSim and ROS2}
Microsoft AirSim is an open-source simulator for autonomous vehicles, developed by Microsoft Research \cite{AirSim_Shah_2017}. It was designed to bridge the gap between simulation and reality in the field of artificial intelligence, particularly for drones and self-driving cars. AirSim provides a platform for researchers and developers to test and train AI algorithms in a realistic, physics-based environment without the risks and costs associated with real-world testing \cite{AirSim_Shah_2017}. The simulator leverages Unreal Engine \cite{UnrealEngine} to create detailed environments and supports various autonomous system sensors like cameras, GPS, and IMUs \cite{CARLA_Dosovitskiy_2017}. To provide communication and integrate external processing, the system supports the Robot Operating System v2 (ROS2)  \cite{ROS2Docs}. AirSim has been upgraded for DARPA by Microsoft and extended with STR Algorithm Development Kit to provide mission generation and randomization.

\section{Problem Formulation} \label{sec:problem}
\begin{figure}[t]
	\centering
	\includegraphics[width=0.49\textwidth]{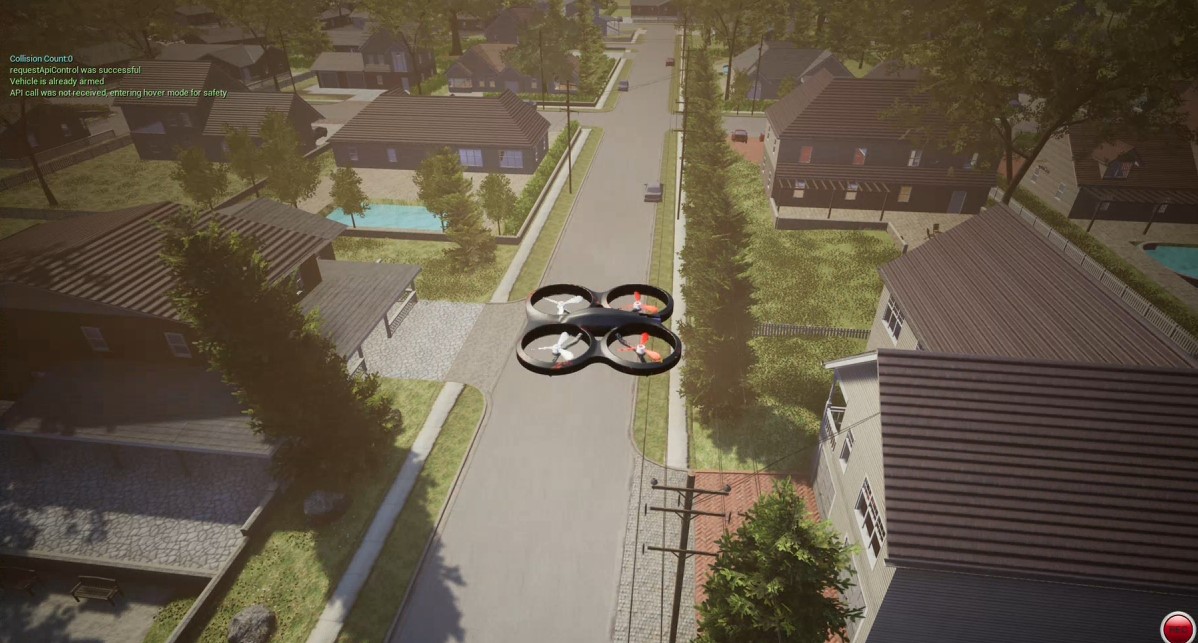}
	\caption{ Our problem features planning the mission of a drone in a neighborhood to search for some targets. The drone is not aware of the real-locations and only have access to the likelihood of targets. }
	\label{fig:AirSim_ROS2}
\end{figure}

Overall, the problem we are interested in is to perform autonomous target localization with multiple targets in an urban environment using an unmanned aerial vehicle (UAV) (\cref{fig:AirSim_ROS2}). A key aspect of this problem is the uncertainty in target locations. The quadrotor does not possess prior knowledge of exact target positions. Instead, it maintains a probabilistic map representing its belief state, which is continuously updated based on a perception system, which eventually leads to update of belief system. The perception system is also responsible for the tracking and identification of the target if they are in the range of the camera of the drone. Note that it is assumed that the quadrotor has only one camera that is pointed in the direction of travel and can only see in a limited area.

Solution to the problem requires perception, belief update, planning and navigation together to generate trajectories that minimize the overall cost function, increases the likelihood of finding targets, avoiding no-fly zones, and minimizing overall flight time.

\subsection{Solution Approach} \label{sec:framework}
\begin{figure}[t]
	\centering
	\includegraphics[width=0.85\columnwidth]{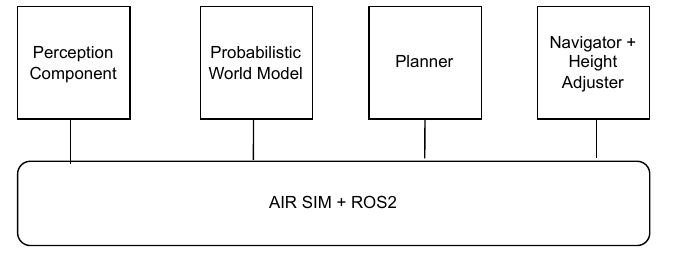}
	\caption{Four key components of our approach. This paper focuses on the planner component.}
	\label{fig:Framework}
\end{figure}

Figure \ref{fig:Framework} shows our software system for mission execution. The perception module detects objects with specific attributes and identifies relationships between them, handling environmental novelties due to varying camera views, occlusion, and weather perturbations. While state-of-the-art object detectors like YOLO world \cite{cheng2024yolo} perform well in closed-world setups, they struggle with novel attributes and relations in SAR operations. Our two-stage approach first detects generic objects, then uses a vision-language model (VLM) \cite{you2023ferret} to detect attributes and relations. VLMs, trained on diverse datasets, can handle novelties effectively.

The probabilistic world model maintains and updates the belief map using raw sensor information received from the perception components, and other sensors such as inertial measurement unit outputting the current drone position, obstacle map, and belief map. It efficiently represents probability distributions and their relations, generating updates from a formal specification derived from mission parameters.  The world model also maintains the flight rules including location of no-fly areas.  The planner component is responsible for generating flight plans -- sequence of waypoints. The navigator component is responsible for finding the shortest and safest path (while avoiding collisions) in the 3-D environment. The height adjustment component is utilized to increase or decrease the height of the drone if no-safe path can be found. Overall, the system works as an interactive protocol between the planner and the navigator. When the planner publishes a waypoint (using ROS2), the determined waypoint is then passed to the navigator component, which guides the quadrotor while ensuring collision avoidance. This process continues until the drone reaches the waypoint or the navigator determines it's unreachable. In either case, a new decision epoch is initiated, allowing the drone to adapt dynamically to its environment based on the most current information.

Note while there are innovations in each of the above component, due to space limitations this paper is restricted to the description of the planning algorithm.

\section{Planner} \label{sec:planner}

In this section, we describe our POMDP framework and our approach for computing near-optimal actions for the POMDP. A major bottleneck in directly using online search algorithms (e.g., POMCP~\cite{POMCP_Silver_2010}) in our setting is scalability---the management of the UAV at each time step can only afford limited latency. To address this problem, we propose a ``shrinking POMCP'' algorithm. Intuitively, once a search tree is constructed, we hypothesize that the agent can traverse the most promising actions down the nodes of the search tree, provided it is in a sparse likelihood region of the map. For example, consider that the agent is in the lower left corner of a grid, with the target likely in the upper left corner. The agent can construct a search tree once and (likely) take multiple steps toward the target region without recomputing again. This technique is designed to dynamically reduce the decision space as planning progresses. The key innovation lies in its ability to guide the agent towards the next best non-sparse region for planning, effectively concentrating computational resources on the most promising areas of the state space. Note that to reduce complexity, we discretize the 3-D state space into a two dimensional slice at a given mission height, set to ensure that the perception component can operate efficiently. If required, the operating height is changed and the 2-D planner can be invoked again. The probabilistic world model can generate the belief distribution at any given height. 

\subsection{POMDP Formalization}
\textbf{Decision Epoch:} In our framework, the decision epoch is defined as the moment when the solver is triggered to determine the next waypoint for the quadrotor. This dynamic decision-making process occurs at discrete time intervals, transitioning from time $t$ to $t+1$, and is initiated by specific events rather than at fixed time intervals. In practice, these events can be monitored and controlled by a meta-controller. Specifically, the solver is activated to make a new decision when one of two conditions is met: either the quadrotor successfully reaches its previously issued waypoint, or it encounters a situation where the current waypoint is unreachable due to obstacles obstructing all valid paths. At each decision epoch, the solver receives a comprehensive update of the system's state, including the most recent belief states, the obstacles detected by the quadrotor's cameras in its immediate vicinity, the quadrotor's current position, and a request for the next waypoint. This event-driven approach to decision epochs ensures that the system remains responsive to the dynamic nature of the environment and the quadrotor's progress, allowing for adaptive and efficient navigation strategies.

\noindent \textbf{States:} In our POMDP framework, we define the state space $\mathcal{S}$ to encompass the position of the quadrotor and the locations of all targets. Let $s_t \in \mathcal{S}$ denote the pre-decision state at time $t$. Each state $s_t$ is represented as a $(3 + 2M)$-dimensional vector:

\begin{equation}
	s_t = [x_q, y_q, z_q, x_1, y_1, \ldots, x_M, y_M]
\end{equation}
where
\begin{itemize}
	\item $(x_q, y_q) \in \mathbb{R}^2$ represents the quadrotor's horizontal position
	\item $z_q \in \mathbb{R}^+$ denotes the quadrotor's altitude
	\item $(x_i, y_i) \in \mathbb{R}^2$ represents the location of the $i$-th target, for $i = \{1, \ldots, M\}$
\end{itemize}

Thus, the complete state space $S$ is of dimensions equalling
$\mathbb{R}^2 \times \mathbb{R}^+ \times (\mathbb{R}^2)^M$. 
This formulation captures the full spatial configuration of the system at any given time $t$, incorporating both the UAV's three-dimensional position and the two-dimensional locations of all $M$ targets within the neighborhood area. To simplify the problem, we discretize the operational area into a grid. Let the original map be a square of side length $L$. We partition this map into an $N \times N$ grid, where each cell represents an area of $(L/N) \times (L/N)$ (in our implementation, $N = 20$, resulting in $20$m $\times 20$m cells). Formally, we can define the grid $G$ as:

\begin{equation}
	G = \{(i, j) \mid i, j \in \{0, 1, \ldots, N-1\}\}
\end{equation}
At each decision epoch, the agent's position is mapped to one of these grid cells.\\

\noindent \textbf{Actions:} 
The action space $\mathcal{A}$ consists of four cardinal directions:

\begin{equation}
	A = \{\text{West}, \text{South}, \text{East}, \text{North}\}
\end{equation}
Each action $a \in \mathcal{A}$ corresponds to moving to an adjacent grid cell in the specified direction. For an agent in cell $(i, j)$ at time $t$, an action $a \in \mathcal{A}$ results in a transition to a new cell $(i', j')$ at time $t+1$, where the new coordinates depend on the chosen direction.

The actual waypoint $w_{t+1}$ within the chosen grid cell is determined by finding a \textit{valid} position closest to the quadrotor's current position $x_t$:

\begin{equation}
	w_{t+1} = \argmin_{w \in V(i',j')} \|w - x_t\|
\end{equation}
where $V(i',j')$ is the set of valid positions within the grid cell $(i', j')$. We describe the translation of the grid-based decision into a specific waypoint for the quadrotor later.\\. 

\noindent \textbf{Observation:} In our POMDP framework, the observation space $\mathcal{O}$ provides partial information about the state. At each time step $t$, an observation $o_t \in \mathcal{O}$ is defined as a tuple $(\gamma_t, B_t)$, where $\gamma_t = (x_q, y_q, z_q) \in \mathbb{R}^3$ represents the exact current position of the quadrotor, and $B_t$ is the updated belief state. The belief state $B_t$ is a probabilistic map over the grid $G$, where for each cell $(i, j) \in G$, $B_t(i, j) \in [0, 1]$ represents the probability of a target being present in that cell.\\

\noindent \textbf{Reward:}
The reward function for our POMDP framework is designed to guide the quadrotor agent in efficiently locating targets within a specified neighborhood area. The reward structure comprises two primary components: target capture and token capture. This dual-component design balances the agent's focus between achieving the primary objective and maintaining comprehensive environmental awareness.

The reward function $R$ is formally defined as:

\begin{equation} \label{eq:reward}
	R = R_\text{target} + \alpha \cdot R_\text{token}
\end{equation}
where $R_\text{target}$ denotes the reward for target capture, $R_\text{token}$ represents the reward for token capture, and $\alpha$ is a hyperparameter controlling the relative importance of token capture.

\textit{Target Capture Reward ($R_\text{target}$):} The target capture component directly addresses the primary objective of the simulation. It is defined as a binary function:

\begin{equation}
	R_\text{target} = 
	\begin{cases}
		1, & \text{if the agent captures a target} \\
		0, & \text{otherwise}
	\end{cases}
\end{equation}

This component provides a significant positive reinforcement upon successful target capture, incentivizing the agent to prioritize navigation towards known or suspected target locations.

\textit{Token Capture Reward ($R_\text{token}$):} The token capture component serves as an exploration incentive, encouraging comprehensive coverage of the environment while prioritizing areas of higher probability. It is defined as:

\begin{equation}
	R_\text{token} = \sum_{i} \mathbbm{1}(i) \cdot P_i
\end{equation}

Where $\mathbbm{1}(i)$ is an indicator function for cell $i$, and $P_i$ is the normalized probability token value for cell $i$. Note that

\begin{equation}
	\mathbbm{1}(i) = 
	\begin{cases}
		1, & \text{if cell $i$ is visited for the first time} \\
		0, & \text{if cell $i$ has been visited before}
	\end{cases}
\end{equation}

and 

\begin{equation}
	P_i = \frac{p_i}{\sum_{j} p_j}
\end{equation}

where $p_i$ represents the raw probability value assigned to cell $i$ in the probabilistic map, and $P_i$ is its normalized form.

This cumulative reward structure incentivizes the exploration of new cells while weighting the reward based on the likelihood of finding targets in each cell. Cells with higher normalized probabilities yield greater rewards upon first visit, potentially facilitating the discovery of targets in areas deemed more promising by the probabilistic map. 

The hyperparameter $\alpha$ in \cref{eq:reward} allows for fine-tuning of the agent's behavior, balancing the emphasis between target acquisition and environmental exploration. A higher $\alpha$ value encourages more thorough exploration, while a lower value prioritizes immediate target capture.

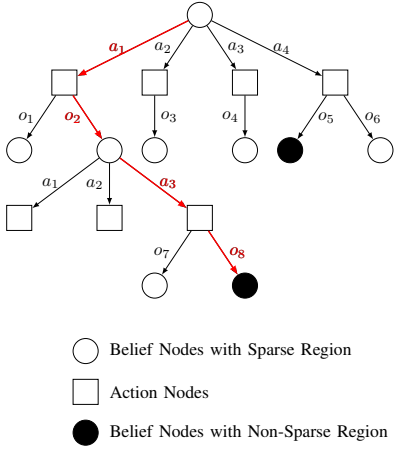
\begin{figure}[t]
\centering
   \resizebox{.6\columnwidth}{!}{
\begin{tikzpicture}[
	scale=0.45,
	level distance=3cm,
	sibling distance=4cm,
	edge from parent/.style={draw, -latex},
	belief node/.style={circle, draw, minimum size=0.5cm},
	belief node 2/.style={circle, draw, fill=black, minimum size=0.5cm},
	action node/.style={rectangle, draw, minimum size=0.5cm},
	]
	\node[belief node] (root) {}
	child {
		node[action node] (a11) {}
		child { 
			node[belief node] {}
			edge from parent node[left] {$o_1$}
		}
		child { 
			node[belief node] (o2) {}
			child{
				node[action node] {}
				edge from parent node[left] {$a_1$}
			}
			child{
				node[action node] {}
				edge from parent node[left] {$a_2$}
			}
			child{
				node[action node] (a13) {}
				child{
					node[belief node] (o7) {}
					edge from parent node[left] {$o_7$}
				}
				child{
					node[belief node 2] (o8) {}
					edge from parent node[right] {$o_8$}
				}
				edge from parent node[right] {$a_3$}
			}
			edge from parent node[left] {$o_2$}
		}
		edge from parent node[left] {$a_1$}
	}
	child {
		node[action node] {}
		child { 
			node[belief node] {} 
			edge from parent node[right] {$o_3$}
		}
		edge from parent node[left] {$a_2$}
	}
	child {
		node[action node] {}
		child { 
			node[belief node] {} 
			edge from parent node[left] {$o_4$}
		}
		edge from parent node[right] {$a_3$}
	}
	child {
		node[action node] {}
		child { 
			node[belief node 2] {} 
			edge from parent node[right] {$o_5$}
		}
		child { 
			node[belief node] (o6) {} 
			edge from parent node[right] {$o_6$}
		}
		edge from parent node[right] {$a_4$}
	}
	;
\draw[red, thick, -latex] (root) -- (a11) node[midway, left] {$a_1$};
\draw[red, thick, -latex] (a11) -- (o2) node[midway, left] {$o_2$};
\draw[red, thick, -latex] (o2) -- (a13) node[midway, right] {$a_3$};
\draw[red, thick, -latex] (a13) -- (o8) node[midway, right] {$o_8$};

\node[below left=1cm and 2cm of o7, anchor=north west, align=left] (legend) {};
\node[belief node, right=0.3cm of legend] (l1) {};
\node[right=0.1cm of l1] {Belief Nodes with Sparse Region};
\node[action node, below=0.3cm of l1] (l2) {};
\node[right=0.1cm of l2] {Action Nodes};
\node[belief node 2, below=0.3cm of l2] (l3) {};
\node[right=0.1cm of l3] {Belief Nodes with Non-Sparse Region};
\end{tikzpicture}
}
\caption{A belief tree constructed by the Shrinking POMCP approach, illustrating its unique decision-making process. Circular nodes represent belief states, with their normalized probability $P(s)$ and quadrotor position $\gamma(s) = (x_q, y_q)$. Black-filled circles indicate non-sparse regions where $P(s) > P_\varepsilon$. Square nodes represent actions. The red arrows show the action sequence $\{a_1, a_2, \ldots, a_k\}$ decided by the agent, where each $a_i = \argmax_a Q(b_i, a)$. This sequence terminates upon reaching either a non-sparse region (black node) or the maximum depth $\text{max\_level}$. Unlike standard MCTS, this approach efficiently guides the agent towards high-probability areas, terminating when $P(b_{k+1}) > P_\varepsilon$ or $k = \text{max\_level}$, thus avoiding goal sampling oscillation.}
\label{fig:belief-action-tree}
\end{figure}

\subsection{Shrinking MCTS} \label{sec:shrinking}
\begin{algorithm}
	\caption{Shrinking POMCP}
	\begin{algorithmic}[1]
		\REQUIRE Initial belief $b_0$, max\_iterations, max\_time, max\_level, $P_\varepsilon$
		\ENSURE Action sequence $A$
		\STATE Initialize belief tree $T$ with root node $b_0$
		\WHILE{iterations $<$ max\_iterations \textbf{and} time $<$ max\_time}
		\STATE $s_0 \sim b_0$
		\STATE SimulateV($s_0$, $b_0$, 0)
		\ENDWHILE
		\RETURN GetActionSequence($b_0$, max\_level, $P_\varepsilon$)
	\end{algorithmic}
	\label{alg:Shrinking_POMCP}
\end{algorithm}

\begin{algorithm}
	\caption{SimulateV($s$, $b$, depth)}
	\begin{algorithmic}[1]
		\IF{IsTerminal($s$) \textbf{or} depth $>$ max\_depth}
		\RETURN 0
		\ENDIF
		\IF{$b$ is leaf node}
		\STATE Expand($b$)
		\RETURN Rollout($s$, $b$)
		\ENDIF
		\STATE $a = \argmax_a \left[Q(b,a) + c \sqrt{\frac{\log(N(b))}{N(b,a)}}\right]$
		\STATE $(s', o, r) = \mathcal{G}(s, a)$
		\IF{$b$ has no child corresponding to $o$}
		\STATE $b' = \text{CreateNewBeliefNode}(b, a, o)$
		\ELSE
		\STATE $b' = b.\text{child}(a, o)$
		\ENDIF
		\STATE $q = r + \gamma \cdot \text{SimulateV}(s', b', \text{depth} + 1)$
		\STATE UpdateStats($b$, $a$, $q$)
		\RETURN $q$
	\end{algorithmic}
	\label{alg:SimulateV}
\end{algorithm}

At each decision epoch, the algorithm constructs a belief tree $T$ with alternating action ($A$) and belief ($B$) nodes. The root node $b_0 \in B$ represents the initial belief state (Line 1 in \cref{alg:Shrinking_POMCP}).  Figure \ref{fig:belief-action-tree} illustrates this tree structure, where action nodes are represented by squares and belief nodes by circles. The red arrows represent the action sequence decided by the agent.

The algorithm begins by sampling a random state $s_0 \sim b_0$ to initialize the root node of the tree (Line 3 in \cref{alg:Shrinking_POMCP}). This initialization step ensures that the search starts from a plausible state within the current belief distribution. The tree expansion process involves a series of iterations, each comprising four main phases: selection, expansion, simulation, and backpropagation. In the selection phase, the algorithm traverses the tree from the root node using the Upper Confidence Bound for Trees (UCT) strategy. For a belief node $b$, the best action $a^*$ is selected according to the equation:
\begin{equation}
	a^* = \argmax_a \left[Q(b,a) + c \sqrt{\frac{\log(N(b))}{N(b,a)}}\right]
\end{equation}
where $Q(b,a)$ is the estimated value of action $a$ in belief state $b$, $N(b)$ is the number of visits to node $b$, $N(b,a)$ is the number of times action $a$ was selected from belief state $b$, and $c$ is an exploration constant (Line 8 in \cref{alg:SimulateV}). This selection strategy balances exploitation of known high-value actions with exploration of less-visited branches.

The simulation phase employs a simulator $\mathcal{G}$, which accepts a state and an action as inputs. This simulator produces three outputs: the probable subsequent state derived from the transition function, the associated observation, and the corresponding reward. This process can be formally represented as:
\[
(s', o, r) \sim \mathcal{G}(s, a)
\]
where $s'$ is the next state, $o$ is the observation, and $r$ is the reward, all generated based on the current state $s$ and action $a$ (Line 9 in \cref{alg:SimulateV}).

The expansion phase occurs when the selected action leads to an unexplored observation. In this case, a new belief node is added to the tree (Line 11 in \cref{alg:SimulateV}). If the observation has been encountered before, the algorithm follows the existing path (Line 13 in \cref{alg:SimulateV}). When a leaf node is reached during the simulation phase, the algorithm expands it by creating child nodes for all possible actions. Then, a rollout is performed till a terminal node. The result of this simulation is then propagated back up the tree in the backpropagation phase, updating the statistics (Q-values and visit counts) of all traversed nodes (Line 16 in \cref{alg:SimulateV}).

\begin{algorithm}
	\caption{GetActionSequence($b$, max\_level, $P_\varepsilon$)}
	\begin{algorithmic}[1]
		\STATE $A = []$, $\text{depth} = 0$
		\WHILE{depth $<$ max\_level \textbf{and} $P(b) \leq P_\varepsilon$}
		\STATE $a = \argmax_a Q(b,a)$
		\STATE $A.\text{append}(a)$
		\STATE $b \sim \mathcal{G}(s, a)$
		\STATE $\text{depth} += 1$
		\ENDWHILE
		\RETURN $A$
	\end{algorithmic}
	\label{alg:action_sequence}
\end{algorithm}

The key innovation of the Shrinking MCTS algorithm lies in its decision-making process, which aims to move the agent to the next best non-sparse region while avoiding goal sampling oscillation. Each belief node $s$ in the tree stores two critical pieces of information: the quadrotor's position $\gamma(s) = (x_q, y_q)$ and the normalized probability $P(s)$ at that position. A belief node is classified as Non-Sparse if its normalized probability exceeds a predefined threshold, i.e., $P(b) > P_\varepsilon$. In \cref{fig:belief-action-tree}, a Belief Node with Non-Sparse Region is represented as a circle filled with black.

Unlike standard MCTS, which typically selects a single action at each step, our Shrinking approach determines an entire action sequence. As shown in \cref{fig:belief-action-tree}, this sequence (represented by red arrows) guides the agent towards high-probability areas. The decision sequence is determined by traversing the tree from the root, selecting the best action at each level until either a maximum depth $\text{max\_level}$ is reached or a Belief Node with Non-Sparse Region is encountered. In the figure, we can see this process leading to the action sequence $\{a_1, a_2, a_3, a_8\}$, terminating at a black node representing a non-sparse region.

Mathematically, this process can be described as taking $k$ actions $\{a_1, a_2, \ldots, a_k\}$ such that $a_i = \argmax_a Q(b_i, a)$ and $b_{i+1} = \tau(b_i, a_i, o_i)$; the constraints we impose for termination are $P(b_{i+1}) \leq P_\varepsilon$ or $k \leq \text{max\_level}$, where $\tau(b, a, o)$ represents the belief update function given action $a$ and observation $o$. The sequence terminates when either $P(b_{k+1}) > P_\varepsilon$ or $k = \text{max\_level}$, as illustrated in \cref{fig:belief-action-tree} where the sequence ends at a black node (non-sparse region). This approach efficiently guides the agent towards promising areas while avoiding the oscillation often seen in standard MCTS implementations.

This approach allows the algorithm to dynamically shrink the decision space by focusing on actions that lead to non-sparse regions, effectively guiding the agent towards areas of the state space with higher certainty. By doing so, the Shrinking POMCP algorithm can potentially overcome the limitations of traditional POMDP solvers in environments with vast or sparse state spaces, leading to more efficient and effective planning in partially observable domains.\\

\noindent \textbf{Rollout}
In Monte Carlo planning, rollouts present a significant computational challenge. These rollouts are essential for approximating the value of leaf nodes within the search tree using a computationally inexpensive strategy. Our approach employs the A* algorithm \cite{A_Star} as the rollout policy, balancing efficiency and effectiveness in pathfinding. While each individual rollout is computationally inexpensive, the cumulative cost of performing thousands of rollouts for each decision becomes a significant bottleneck. To address this challenge and improve overall performance, we implement the A* algorithm using C programming language \cite{C_Programming}. We found that this approach led to better results than standard random rollouts.

Each belief node $s$ in our tree structure encapsulates information about the quadrotor's position $p(s)$. This positional information is represented at two distinct levels of granularity:
\begin{enumerate}
    \item Fine-grain level: The exact position on a high-resolution map of dimensions $L \times L$.
    \item Discretized level: A coarser $N \times N$ grid, where each cell corresponds to a region of the fine-grain map.
\end{enumerate}

While the Monte Carlo Tree Search (MCTS) simulation operates on the discretized $N \times N$ grid for computational efficiency, the rollout process necessitates a more precise position determination. Given a current state and action, we first identify the target cell in the discretized grid. The challenge then becomes determining the exact position within this target cell.

To address this, we employ a sampling-based approach that balances accuracy and computational efficiency. Let $\gamma_c$ represent the set of all possible positions in the current cell, and $\gamma_t$ represent the set of all possible positions in the target cell. We define a sampling function $S(\gamma_t)$ that returns a subset of positions from $\gamma_t$.

From this sampled subset, we identify the set of valid positions $V(S(\gamma_t))$:
\begin{equation}
	V(S(\gamma_t)) = \{\gamma \in S(\gamma_t) \mid \gamma \text{ is a valid position}\}
\end{equation}
The next exact position $\gamma_{next}$ is then determined by finding the position in $V(S(\gamma_t))$ that minimizes the Euclidean distance from the current position $\gamma_{current}$:

\begin{equation}
	\gamma_{next} = \argmin_{\gamma \in V(S(\gamma_t))} ||\gamma - \gamma_{current}||
\end{equation}
where $||\cdot||$ denotes the Euclidean distance. 

Once $\gamma_{next}$ is determined, we calculate the rollout value by running the A* algorithm from $\gamma_{current}$ to $\gamma_{next}$, using the given obstacle map. Let $L(\gamma_{current}, \gamma_{next})$ denote the path length returned by A*. The rollout value $R$ is then computed as a function of this path length:

\begin{equation}
	R = f(L(\gamma_{current}, \gamma_{next}))
\end{equation}
where $f$ is a monotonically decreasing function. This formulation ensures that shorter paths, indicating easier navigation, result in higher rollout values, while longer paths, suggesting more complex navigation, yield lower values.

\subsection{Height Adjustment} \label{sec:height}
Our planning approach initially assumes constant altitude but incorporates an adaptive height adjustment strategy to balance obstacle avoidance with smooth flight patterns, mitigating undesirable "up and down" motions due to system noise.

The quadrotor starts at altitude $h_{init}$. For each target cell $C_t$ in the planned path, we evaluate the number of valid positions $N_v(C_t, h)$ at current height $h$:

\begin{equation}
    N_v(C_t, h) = |\{\gamma \in C_t \mid \gamma \text{ is valid at height } h\}|
\end{equation}

We define an ``obstacle\_tolerance\_threshold'' $\tau$. If $N_v(C_t, h) \geq \tau$, the waypoint's altitude remains unchanged. Otherwise, we incrementally adjust the height:

\begin{equation}
    h_{new} = \min(h + \Delta h, h_{max})
\end{equation}

where $\Delta h$ is the height increment and $h_{max}$ is the maximum allowed height. This process repeats until $N_v(C_t, h_{new}) \geq \tau$ or $h_{new} = h_{max}$.

If $N_v(C_t, h_{max}) < \tau$, we designate the cell as a no-fly zone and re-execute the Monte Carlo Tree Search (MCTS) algorithm for an alternative path. This strategy ensures safe obstacle avoidance while minimizing unnecessary altitude changes, resulting in smoother and more efficient flight trajectories.

\section{Analysis and Evaluation} \label{sec:analysis}
We evaluate our proposed framework using both the AirSim-ROS2 simulator and the two-dimensional simulator described in \cref{sec:framework} and report the performance of Shrinking POMCP against baseline.

\subsection{Environment Simulators} \label{sec:environment_simulators}
We test our approach in two simulation environments:

\begin{enumerate}
    \item \textbf{3D AirSim-ROS2 Simulator:} This advanced environment incorporates the full functionality of our framework, including the Probabilistic World Model and Navigator components.
    
    \item \textbf{Two-dimensional Simulator:} In this simplified environment, we test our approach without considering altitude effects. The Probabilistic World Model and the Navigator are replaced with surrogate ROS2 nodes that provide equivalent functionality, allowing for efficient testing and validation of our algorithms.
\end{enumerate}

\subsection{Hyperparameters} \label{Hyperparameters}
The testing environment for this mission scenario is set within a 400m $\times$ 400m map, with a strict time constraint of 5 minutes to locate all targets. The quadrotor's initial altitude is set at 10 meters, providing a balance between coverage area and detail resolution.

Several hyperparameters are adjusted to optimize the search strategy. The discount factor in the POMDP framework is tested with values of 0.8, 0.9, and 0.995, influencing the balance between immediate and future rewards. The $\alpha$ value in the reward function discussed in \cref{sec:planner}, which affects the weighting of different reward components, is varied among 0, 1, and 10. Shrinking MCTS runs 3000 iterations for every decision epoch, with the exploration parameter in UCT set to $\sqrt{2}$. For vertical navigation, the height adjustment parameter $\delta h$ (discussed in \cref{sec:height}) is set to 3m, with a maximum allowable altitude of 30m. These parameters collectively define the operational constraints and decision-making framework for the quadrotor's search mission.

\vspace{-1cm}
\subsection{Baselines} \label{sec:baselines}
In our experiments, we evaluate the performance of our Shrinking POMCP approach against three baseline methods: Lawnmower algorithm, Greedy algorithm, and standard MCTS (POMCP without shrinking). We test these methods on two types of belief maps: Uniform belief map and Sparse belief map with one peak. All methods are compared on the same map, with identical no-fly zones and starting agent positions for each scenario to ensure a fair comparison.

\textbf{Lawnmower Algorithm:} This baseline performs a systematic search in non-zero belief areas. At the start of each episode, the agent moves to the nearest non-zero probability position on the probability map. It then executes a lawnmower pattern search, systematically covering the non-zero belief region until all targets are found. This method ensures complete coverage of the search area but may not be optimal in terms of efficiency.

\textbf{Greedy Algorithm:} This approach makes local decisions based on immediate information. At each step, the agent selects its next position by choosing the adjacent cell with the highest value in the resized array. While this method can be effective in quickly identifying high-probability areas, it may suffer from getting trapped in local maxima.

\textbf{Monte Carlo Tree Search (MCTS):} We implement the standard POMCP algorithm without our proposed shrinking approach as a baseline. This method makes decisions only for cardinal directions (WEST, SOUTH, EAST, NORTH), moving to the next adjacent cell in the discretized grid at each step. This baseline allows us to directly compare the performance gains achieved by our Shrinking POMCP approach.

\subsection{Results (Two-Dimensional)} \label{sec:2D_results}
\begin{figure}[t]
	\centering
	\includegraphics[width=0.45\textwidth]{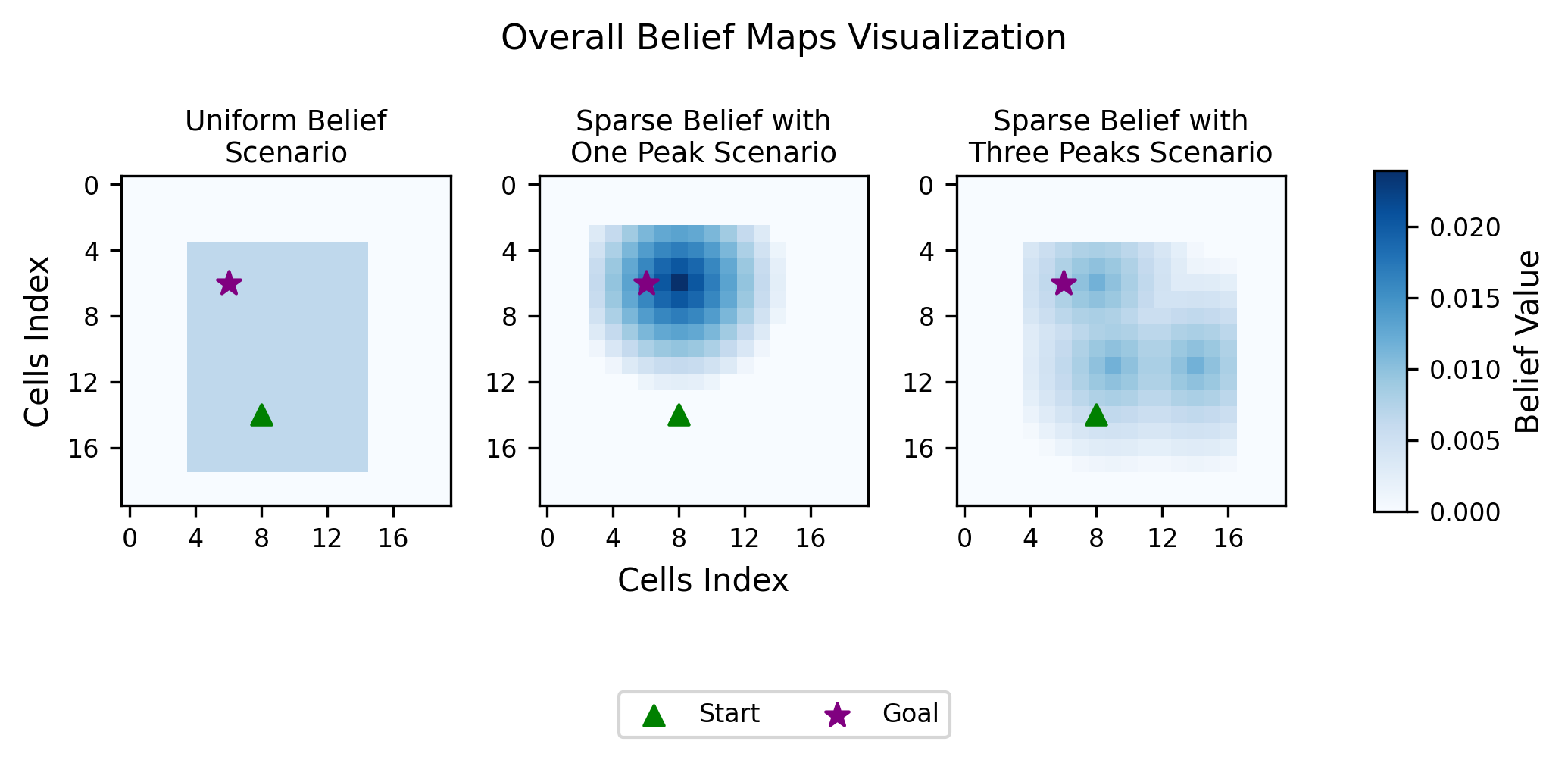}
	\caption{Environments for different belief scenarios. Left: Uniform belief distribution across the environment. Center: Sparse belief with a single peak, indicating high certainty in one area. Right: Sparse belief with three peaks, representing multiple areas of high certainty. The green triangle ($\blacktriangle$) represents the start position, and the purple star ($\star$) indicates the goal position. Color intensity corresponds to belief value, with darker blue indicating higher belief.}
	\label{fig:combined_belief_maps}
\end{figure}

\begin{table}[htbp]
\centering
\caption{Hyperparameter Sweep Results: DF represents discount factor, RA is reward alpha. SE is standard error of mean. Bold numbers represent the optimal combination of discount factor and reward alpha chosen for each belief type, based on the best average performance.}
\label{tab:hyperparameter_sweep}
\small
\begin{tabular}{ccccccccc}
\hline
\multirow{2}{*}{\textbf{DF}} & \multirow{2}{*}{\textbf{RA}} & \multicolumn{2}{c}{\textbf{Uniform Belief}} & \multicolumn{2}{c}{\textbf{One Peak}} & \multicolumn{2}{c}{\textbf{Three Peaks}} \\
\cline{3-8}
& & Mean & SE & Mean & SE & Mean & SE \\
\hline
0.8 & 0 & 17.3 & 3.4 & 17.2 & 5.1 & 6.2 & 1.0 \\
0.8 & 1 & 24.3 & 3.4 & 15.2 & 4.8 & 8.4 & 3.5 \\
0.8 & 10 & 37.4 & 14.5 & 12.5 & 4.8 & 12.6 & 6.0 \\
0.9 & 0 & 18.1 & 9.9 & 15.9 & 5.1 & 6.1 & 0.9 \\
0.9 & 1 & 16.1 & 7.7 & 12.2 & 5.0 & 15.4 & 6.3 \\
0.9 & 10 & 15.3 & 8.4 & 12.4 & 4.9 & 6.7 & 3.3 \\
0.995 & 0 & \textbf{5.7} & \textbf{1.6} & 12.0 & 4.9 & 11.3 & 4.7 \\
0.995 & 1 & 7.3 & 1.6 & 12.2 & 5.0 & 6.7 & 2.0 \\
0.995 & 10 & 22.7 & 9.7 & \textbf{11.3} & \textbf{4.8} & \textbf{3.0} & \textbf{0.9} \\
\hline
\end{tabular}
\vspace{0.5cm}

\end{table}
\textbf{Belief Maps:} Three types of initial belief distributions are tested in this scenario. The first is a Uniform Belief (\cref{fig:combined_belief_maps}), where probabilities are evenly distributed across the entire map. The second is a Sparse Belief with One Peak (\cref{fig:combined_belief_maps}), characterized by a high probability concentration at a single location that gradually diffuses outward. The third is a Sparse Belief with Three Peaks (\cref{fig:combined_belief_maps}), featuring multiple areas of high probability concentration. In all scenarios, the belief distribution changes to reflect different levels of prior knowledge about the environment.

\textbf{Hyperparameter Sweep:} We conducted a comprehensive hyperparameter sweep to evaluate the performance of our framework under different conditions and compare it with baseline algorithms. The experiments were performed on the three types of belief maps:

\textbf{Uniform Belief Scenario:} For the Uniform Belief scenario as shown in \cref{tab:hyperparameter_sweep}, we explored different combinations of discount factors (0.8, 0.9, 0.995) and reward alphas (0, 1, 10). The performance metric is the number of decision epochs required to locate all targets, with a maximum limit of 100 epochs. The results indicate that higher discount factors generally lead to better performance. This suggests that in uniformly distributed belief scenarios, our framework benefits from considering long-term rewards more heavily.

\textbf{Sparse Belief with One Peak Scenario:} In the Sparse Belief with One Peak scenario as shown in \cref{tab:hyperparameter_sweep}, we used a similar experimental setup. Similar as Uniform Belief, higher discount factors generally lead to better performance.

\textbf{Sparse Belief with Three Peaks Scenario:} For the Sparse Belief with Three Peaks scenario, higher discount factors, particularly when combined with higher reward alpha values, indicate better performance for our approach. This suggests that in more complex belief distributions with multiple high-probability areas, our framework benefits from both long-term planning (higher discount factors) and a stronger emphasis on immediate rewards (higher reward alpha).

Based on the results of our hyperparameter sweep, we selected the combination of discount factor and reward alpha that achieved the best average performance for each belief type. These optimal hyperparameter values were then used consistently in both our 2D environment and AirSim environment experiments, ensuring a standardized approach across different simulation platforms.

\begin{figure}[ht!]
	\centering
	\includegraphics[width=0.49\textwidth]{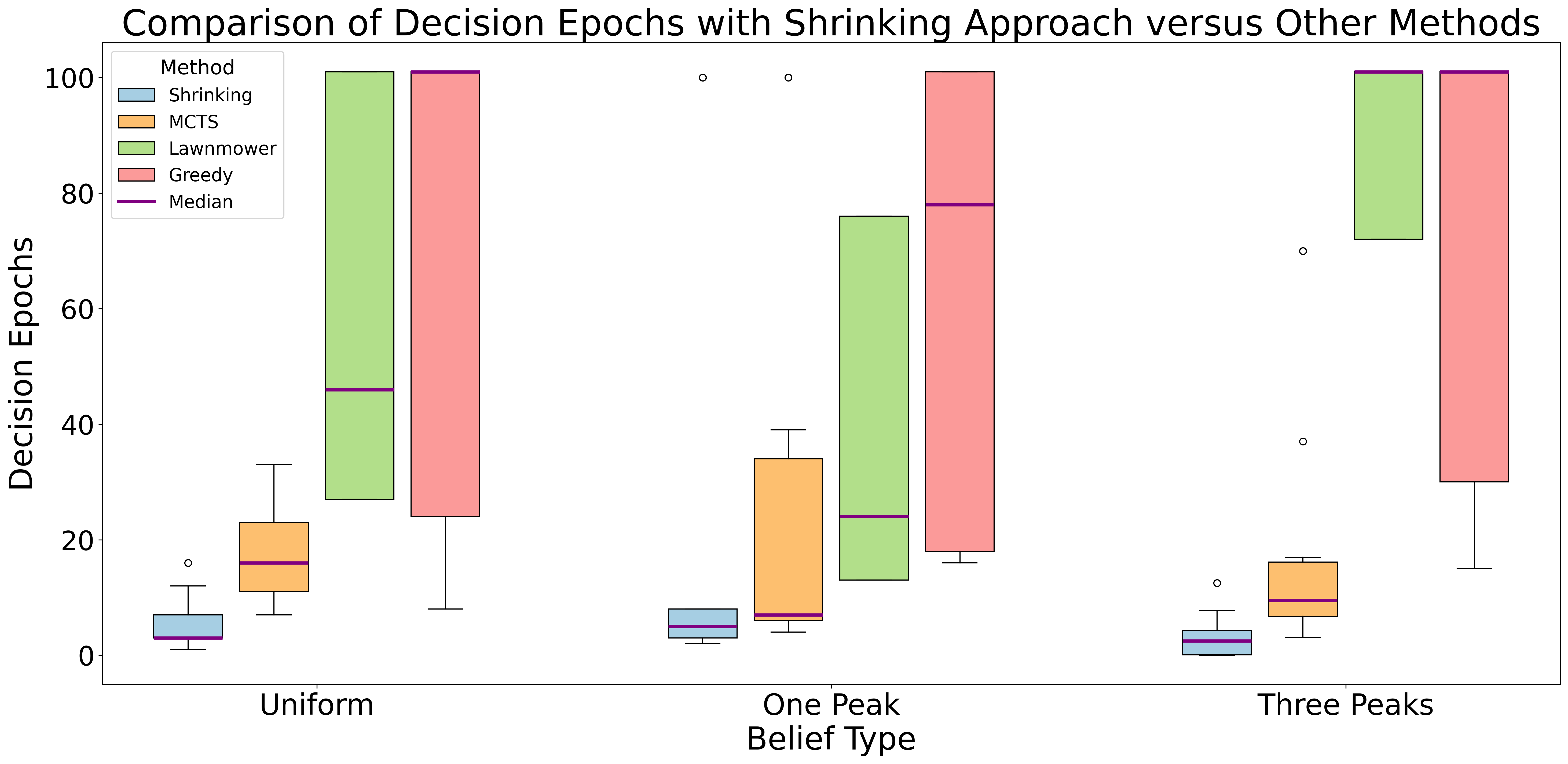}
	\caption{Comparison between our Shrinking approach and other methods (MCTS, Lawnmower, and Greedy algorithms). Shrinking POMCP requires significantly fewer decision epochs to locate all targets across all belief types. Colors represent different approaches: Blue - Shrinking, Orange - MCTS, Green - Lawnmower, Red - Greedy.}
	\label{fig:shrinking}
\end{figure}

\textbf{Comparison between Shrinking POMCP and Other Methods:} The performance comparison between our Shrinking approach and other methods (MCTS, Lawnmower, and Greedy algorithms) demonstrates the superior efficiency of our proposed method. As shown in Figure \ref{fig:shrinking}, the Shrinking approach consistently requires significantly fewer decision epochs to locate all targets across both Uniform and One Peak belief types.

The key advantage of our Shrinking POMCP lies in its ability to output an action sequence at each decision epoch, rather than a single action. This enables the agent to efficiently navigate to the next best non-sparse region for planning in every epoch. By doing so, our approach effectively mitigates a common challenge faced by traditional POMCP, where sampling goals can cause the agent to oscillate between different actions. The result is a more decisive and efficient search strategy, as evidenced by the consistently lower number of decision epochs required across different belief scenarios.

\subsection{Results (AirSim-ROS2)} \label{sec:airsim_results}

\begin{figure}[t]
	\centering
	\includegraphics[width=0.43\textwidth]{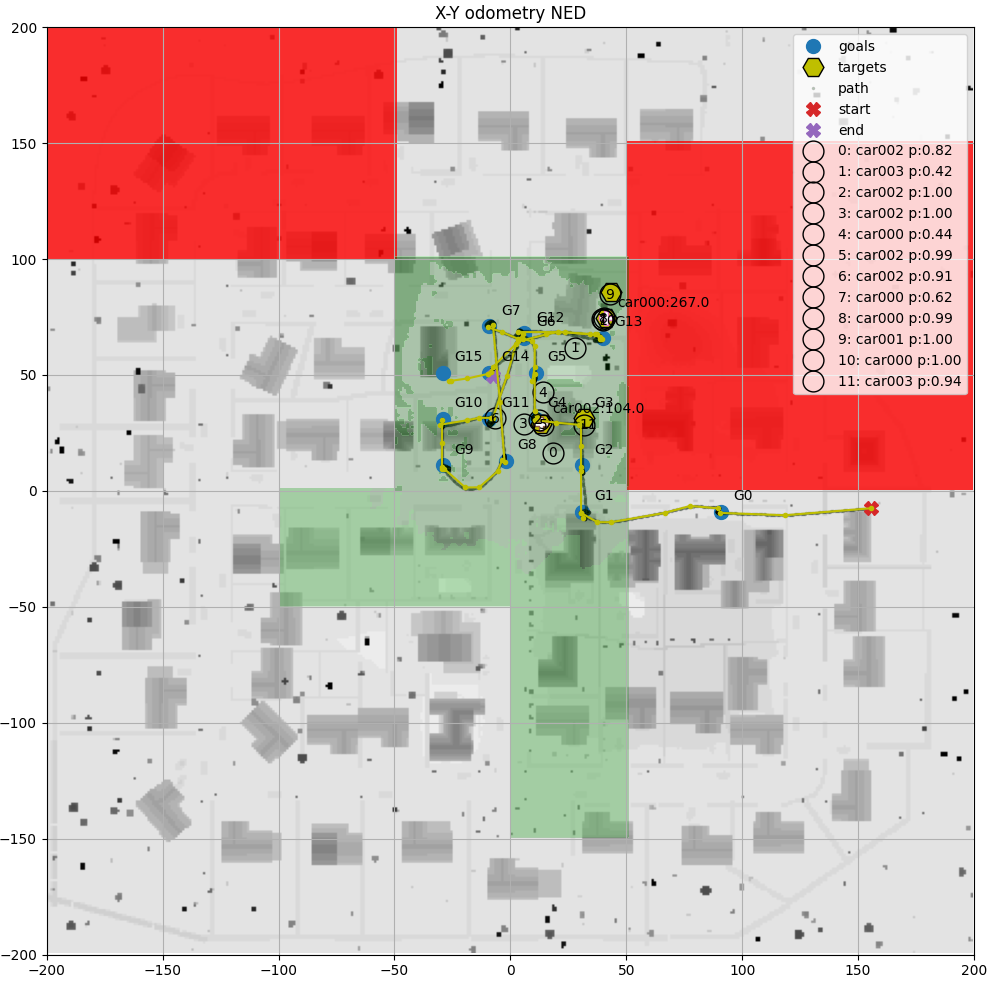}
	\caption{X-Y odometry plot (in North East Down coordinates) of a quadrotor drone's area search mission, showing the drone's path (yellow), \textcolor{black}{ground truth targets (hexagons), detected targets (numbered circles)}, no-fly zones (red), buildings (grey), and search areas (green). The legend lists detected targets with their probabilities.}
	\label{fig:path_plot}
\end{figure}

\textbf{No-Fly-Zones:} No-fly zones provide spatial and temporal constraints for quadrotor operations in simulated environments. These zones are defined areas that the quadrotor should not enter. Each no-fly zone is represented by geometric boundaries, which could be 3D shapes like cubes, cylinders, or complex polygons. A key feature of AirSim's no-fly zones is their temporal aspect. Each zone includes \texttt{no\_earlier\_than} and \texttt{no\_later\_than} parameters, specifying the time window during which the restriction is active.

\textbf{Entities of Interest:} The configuration defines specific vehicles as targets, each with unique attributes (e.g., a red SEDAN). As the quadrotor explores, its cameras identify vehicles matching these criteria. The system processes captured images to update a belief map, reflecting the likelihood of finding target entities in different locations.

\textbf{Evaluation Scenarios:} We evaluate the components by running them through simulations in the mission scenario (example of one scenario shown in  figure \ref{fig:path_plot}). A \textbf{scenario} is defined as a single mission and environmental configuration, consisting of mission-related aspects (e.g., target types and arrangements, areas of interest, keep-out zones, and belief maps) and environmental factors (e.g., weather, time of day, and camera noise). 

\textbf{Metrics:} Overall, the success of the mission is evaluated using a set of metrics: 
\textbf{COP Completeness\footnote{COP stands for common operating picture}:}
Percentage of correctly reported target information elements out of total target information elements. It's calculated by maintaining a running status of each information element (e.g., type, location) throughout the trial, determining if it's in scope and correctly reflected in the COP. The metric is presented as a distribution with mean and 95 \% CI for each evaluation condition. 
\textbf{COP Accuracy:}
Percentage of correctly reported targets out of total reports. This metric is captured per trial and presented as a distribution with mean and 95 \% CI for each evaluation condition, reflecting the accuracy of the system's target identification and reporting; and
\textbf{COP Reporting Latency:}
Average time between a target change and its correct reporting. It's calculated using time points for every relevant change during each trial at a 1-second resolution. The metric is presented as a distribution with mean and 95\% CI for each evaluation condition, indicating the system's responsiveness to target changes.

The figure (\cref{fig:path_plot})
describes a single run of the system for one of the scenarios. 
The map is divided into different zones: green areas represent regions \textcolor{black}{of interest} with non-zero probability of containing targets, grey areas indicate \textcolor{black}{obstacles, like buildings and trees}, and red areas denote no-fly zones. The drone's path, shown in yellow, navigates efficiently through the green areas, avoiding obstacles and restricted zones. In this run, our planner algorithm 
located all \textcolor{black}{4} cars of interest, marked as numbered targets on the map \textcolor{black}{using the novel perception component developed by SRI in the same DARPA project}. The drone's path demonstrates an efficient search strategy, entering high-probability areas directly and minimizing time spent in low-probability regions. This efficiency is  due to our shrinking approach in the algorithm, allowing the quadrotor to make faster decisions and focus on promising areas.

\begin{figure}[t]
	\centering
	\includegraphics[width=0.4\textwidth]{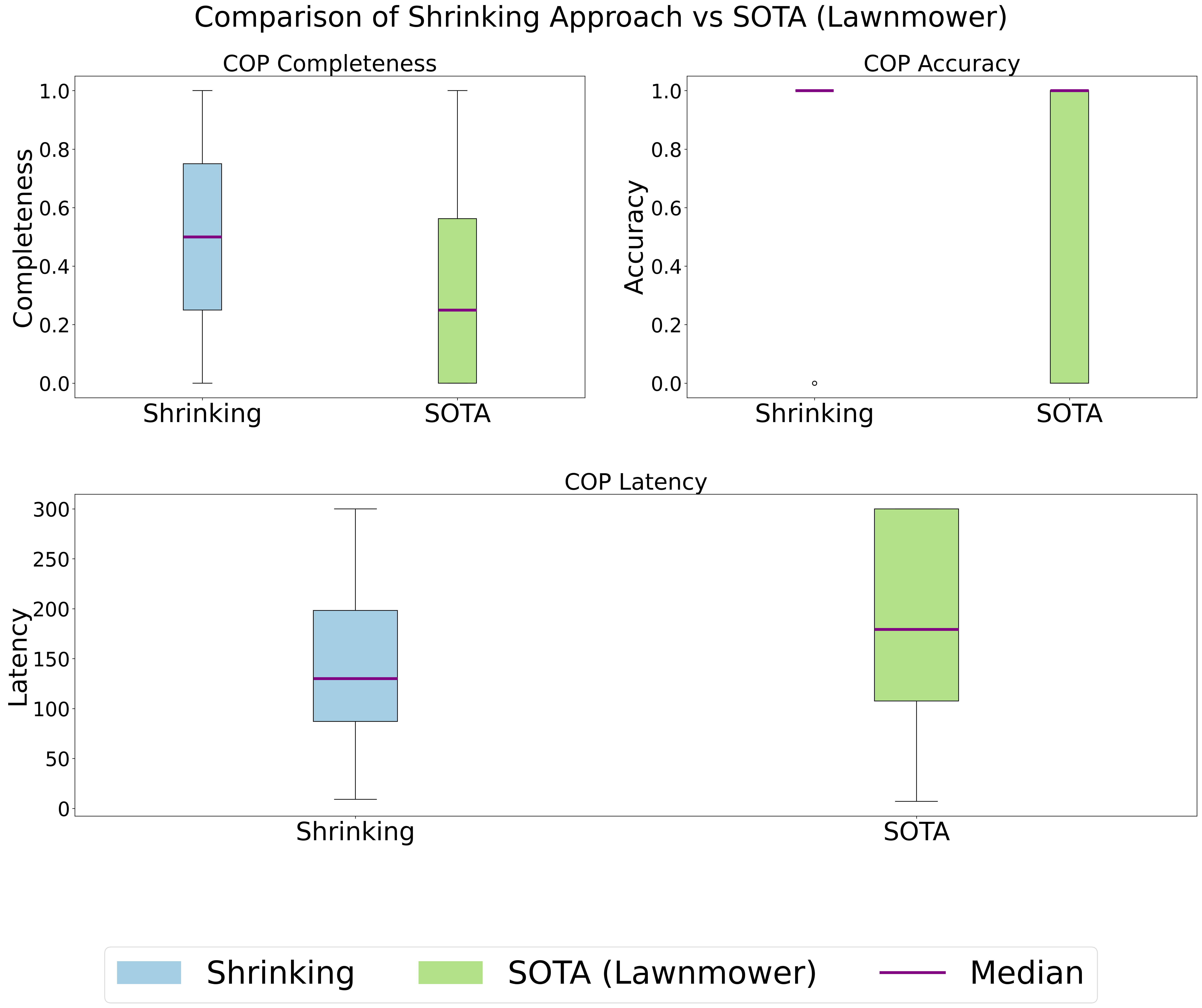}
	\caption{Performance comparison between the Shrinking POMCP approach and SOTA (Lawnmower) method. The box plots show the distribution of COP Completeness, COP Accuracy, and COP Latency across multiple runs. The Shrinking approach (light blue) consistently outperforms the SOTA method (light green) across all metrics, with higher completeness and accuracy, and lower latency. Median values are indicated by purple lines. }
	\label{fig:cop}
\end{figure}

Figure \ref{fig:cop} illustrates the performance of our Shrinking POMCP framework compared to the \textcolor{black}{SOTA Baseline (lawnmover search pattern with YOLO v8 World perception)} across multiple scenarios and runs. We show variance to remove experimental bias. 
Our method consistently outperforms the Baseline in all three key metrics: COP Completeness, COP Accuracy, and COP Latency. For COP Completeness, our approach shows a higher median and a more concentrated distribution, indicating more consistent and thorough coverage. In terms of COP Accuracy, our Shrinking approach demonstrates superior performance, with a median accuracy of 1.0 compared to the SOTA's lower and more varied distribution. The average COP Accuracy for our approach is \textbf{0.81}, while the SOTA achieves only \textbf{0.57}, highlighting our method's significantly higher precision in correctly reporting targets. Regarding COP Latency, our framework exhibits lower latency with a tighter distribution, suggesting faster and more consistent response times. These enhancements are due to our framework's efficient search strategy, enabling the quadrotor to make faster decisions and prioritize high-probability areas.

\section{Conclusion} \label{sec:conclusion}
In this paper, we presented an optimized approach for UAV-based search and rescue operations in neighborhood areas. We developed a realistic simulator using AirSim and ROS2, and formulated the path planning problem as a POMDP. Our Shrinking POMCP approach addresses time constraints in SAR missions. Experimental results demonstrate that this method significantly outperforms alternatives, locating all targets in fewer decision epochs. This suggests our solution can enhance the efficiency of UAV-assisted SAR missions, saving critical time in emergencies.

\section{Discussion and Future Work} \label{sec:future_work}
While Shrinking POMCP shows promise in adapting to non-stationary environments, future research could explore integrating function approximation techniques into the planning process. Neural network approximators could be used to learn the inherent uncertainty in the environment \cite{ADA-MCTS}, potentially improving the algorithm's ability to adapt to changes. Additionally, leveraging learned approximators to accelerate MCTS convergence \cite{PAMCTS} could enhance computational efficiency. Combining these approaches with our Shrinking POMCP could lead to a more robust and efficient algorithm capable of rapid adaptation in non-stationary environments while maintaining computational tractability, particularly in domains with large state spaces and complex dynamics.

\section*{Acknowledgments}
This material is based upon work sponsored in part by the by the Defense Advanced Research Projects Agency (DARPA) \textcolor{black}{under its Assured Neuro Symbolic Learning and Reasoning (ANSR) project and the US Air Force Research Laboratory (AFRL)} and in part by the National Science Foundation Grant 2238815. Any opinions, findings, and conclusions or recommendations expressed in this material are those of the authors and do not necessarily reflect the views of DARPA. The authors would like to thank Microsoft for creating the AirSim simulator and STR for developing the challenge scenarios for the ANSR program.

\vspace{-0.2cm}
\bibliographystyle{ieeetr}
\bibliography{references} 

\end{document}